\documentclass{article}
\usepackage[nonatbib,final]{nips_2016}
\usepackage[numbers]{natbib}
\usepackage[utf8]{inputenc}
\usepackage[T1]{fontenc}
\usepackage{booktabs}
\usepackage{nicefrac}
\usepackage{microtype}
\usepackage{amsmath}
\usepackage{amsthm}
\usepackage{amssymb}
\usepackage{amsfonts}
\usepackage{bm}
\usepackage{graphicx}
\usepackage{subfigure} 
\usepackage{makecell}
\usepackage{multirow}
\usepackage{hyperref}
\usepackage{url}

\title{Transitive Hashing Network for Heterogeneous Multimedia Retrieval}
\author{
Zhangjie Cao$^\dag$, Mingsheng Long$^\dag$, and Qiang Yang$^\ddag$\\
$^\dag$School of Software, Tsinghua University, Beijing 100084, China\\
$^\ddag$Hong Kong University of Science and Technology, Hong Kong\\
\texttt{caozhangjie14@gmail.com, mingsheng@tsinghua.edu.cn, qyang@cse.ust.hk}\\
}

\begin{document}

\maketitle

\begin{abstract}
Hashing has been widely applied to large-scale multimedia retrieval due to the storage and retrieval efficiency. Cross-modal hashing enables efficient retrieval from database of one modality in response to a query of another modality. Existing work on cross-modal hashing assumes heterogeneous relationship across modalities for hash function learning. In this paper, we relax the strong assumption by only requiring such heterogeneous relationship in an auxiliary dataset different from the query/database domain. We craft a hybrid deep architecture to simultaneously learn the cross-modal correlation from the auxiliary dataset, and align the dataset distributions between the auxiliary dataset and the query/database domain, which generates transitive hash codes for heterogeneous multimedia retrieval. Extensive experiments exhibit that the proposed approach yields state of the art multimedia retrieval performance on public datasets, i.e. NUS-WIDE, ImageNet-YahooQA.
\end{abstract}

\section{Introduction}
Multimedia retrieval has attracted increasing attention in the presence of multimedia big data emerging in search engines and social networks. Cross-modal retrieval is an important paradigm of multimedia retrieval, which supports similarity retrieval across different modalities, e.g. retrieval of relevant images with text queries. A promising solution to cross-modal retrieval is hashing methods, which compress high-dimensional data into compact binary codes and generate similar codes for similar objects \cite{cite:Arxiv14Hashing}. To date, effective and efficient cross-modal hashing remains a challenge, due to the heterogeneity across  modalities \cite{cite:KDD14HTH}, and the semantic gap between features and semantics \cite{cite:TPAMI00SemanticGap}.

\begin{figure}[tbp]
\centering
\includegraphics[width=1.0\columnwidth]{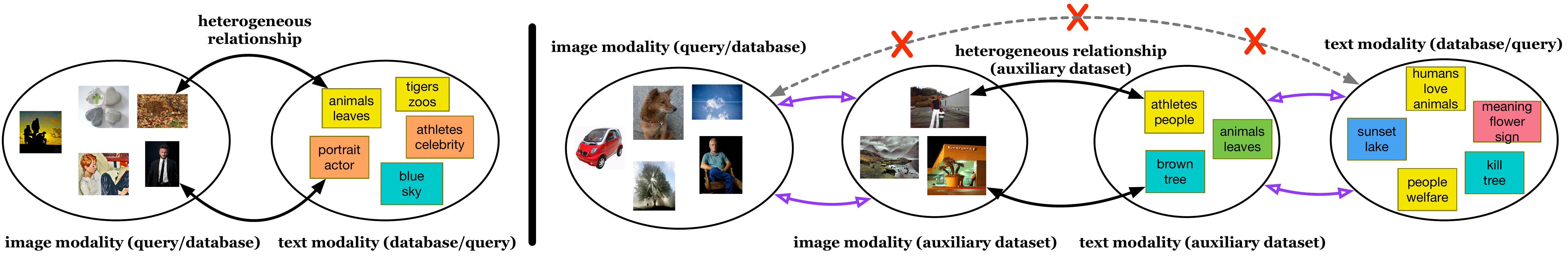}
\caption{Problem overview. (left) Traditional cross-modal hashing, where heterogeneous relationship between query and database  (black arrows) is available for hash learning. (right) The new transitive hashing, where heterogeneous relationship is not directly available between query and database (dashed arrows) but is available from an auxiliary dataset of different distributions (purple arrows).}
\label{fig:problem}
\end{figure}

An overview of cross-modal retrieval problems is shown in Figure~\ref{fig:problem}. Traditional cross-modal hashing methods \cite{cite:CVPR10CMSSH,cite:IJCAI11CVH,cite:NIPS12CRH,cite:SIGMOD13IMH,cite:PAMI14CMNN,cite:AAAI14SCM,cite:IJCAI15QCH,cite:JDCMH16} have achieved promising performance for multimedia retrieval. However, they all require that the heterogeneous relationship between query and database is available for hash function learning. This is a very strong requirement for many practical applications, where such heterogeneous relationship is not available. For example, a user of YahooQA (Yahoo Questions and Answers) may hope to search images relevant to his QAs from an online social media such as ImageNet. Unfortunately, since there are no link connections between YahooQA and ImageNet, it is not easy to satisfy the user's information need. Therefore, how to support cross-modal retrieval without direct relationship between query and database is an interesting problem worth investigation.

This paper proposes a novel transitive hashing network (THN) approach to address the above problem, which generates compact hash codes of images and texts in an end-to-end deep learning architecture to construct the transitivity between query and database of different modalities. As learning cross-modal correlation is impossible without any heterogeneous relationship information, we leverage an auxiliary dataset readily available from a different but related domain (such as Flickr.com), which contains the heterogeneous relationship (e.g. images and their associated texts). We craft a hybrid deep network to enable  heterogeneous relationship learning on this auxiliary dataset. Note that, the auxiliary dataset and the query/database sets are collected from different domains and follow different data distributions, hence there is substantial dataset shift which poses a major difficulty to bridge them. To this end, we further integrate a homogeneous distribution alignment module to the hybrid deep network, which closes the gap between the auxiliary dataset and the query/database sets. Based on heterogeneous relationship learning and homogeneous distribution alignment, we can construct the transitivity between query and database in an end-to-end deep architecture to enable efficient heterogeneous multimedia retrieval. Extensive experiments show that our THN model yields state of the art multimedia retrieval performance on public datasets, i.e. NUS-WIDE, ImageNet-YahooQA.

\section{Related Work}
This work is related to hashing for multimedia retrieval, known as cross-modal hashing, which has been an increasingly popular research topic in machine learning, computer vision, and multimedia retrieval communities \cite{cite:CVPR10CMSSH,cite:IJCAI11CVH,cite:NIPS12CRH,cite:SIGMOD13IMH,cite:PAMI14CMNN,cite:AAAI14SCM,cite:IJCAI15QCH,cite:JDCMH16,cite:KDD16DVSH}. Please refer to \cite{cite:Arxiv14Hashing} for a comprehensive survey.

Previous cross-modal hashing methods can be organized into unsupervised methods and supervised methods. Unsupervised methods learn hash functions that encode input data points to binary codes only using unlabeled training data. Typical learning criteria include reconstruction error minimization \cite{cite:VLDB14MSAE}, neighborhood preserving in graph-based hashing \cite{cite:IJCAI11CVH,cite:SIGMOD13IMH}, and quantization error minimization in correlation quantization \cite{cite:IJCAI15QCH,cite:SIGIR16CCQ}. Supervised methods explore supervised information (e.g. pairwise similarity or relevance feedback) to learn compact hash codes. Typical learning criteria include metric learning \cite{cite:CVPR10CMSSH}, neural network \cite{cite:PAMI14CMNN}, and correlation learning \cite{cite:AAAI14SCM,cite:IJCAI15QCH}. As supervised methods can explore the semantic relationship to bridge modalities  and reduce the semantic gap \cite{cite:TPAMI00SemanticGap}, they can achieve superior accuracy than unsupervised methods for cross-modal retrieval. 

Most of previous cross-modal hashing methods based on shallow architectures cannot effectively exploit the heterogeneous relationship across different modalities. Latest deep models for multimodal embedding \cite{cite:NIPS13Devise,cite:NIPS14MNLM,cite:CVPR15LRCN,cite:NIPS15mQA} have shown that deep learning can bridge heterogeneous modalities more effectively for image captioning, but it remains unclear how to explore these deep models to cross-modal hashing. Recent deep hashing methods \cite{cite:AAAI14CNNH,cite:CVPR15DNNH,cite:AAAI16DHN} have given state of the art results on many datasets, but they can only be used for single-modal retrieval. To the best of our knowledge, DCMH \cite{cite:JDCMH16} is the only cross-modal deep hashing method that uses deep convolutional networks \cite{cite:NIPS12CNN} for image representation and multilayer perceptrons \cite{cite:MIT86MLP} for text representation. However, DCMH can only address traditional cross-modal retrieval where heterogeneous relationship between query and database is available for hash learning, which is very restricted for real applications. To this end, we propose a novel transitive hashing network (THN) method to address cross-modal retrieval where heterogeneous relationship is not available between query and database, which leverages an auxiliary cross-modal dataset from a different domain and builds transitivity to bridge different modalities.

\section{Transitive Hashing Network}
In transitive hashing, we are given a query set ${\cal X}^q = \{ {\bf x}_i\}_{i=1}^n$ from modality $X$ (such as image), and a database set ${\cal Y}^d = \{ {\bf y}_j\}_{j=1}^m$ from modality $Y$ (such as text), where ${\bf x}_i \in \mathcal{\bf{R}}^{d_x}$ is a $d_x$-dimensional feature vector in the query modality and ${\bf y}_j \in \mathcal{\bf{R}}^{d_y}$ is a $d_y$-dimensional feature vector in the database modality. A key challenge of transitive hashing is that no supervised relationship  is available between query and database. Therefore, we bridge modalities $X$ and $Y$ by learning from an auxiliary dataset ${\cal\bar{X}} = \{ {\bar{\bf x}}_i\}_{i=1}^{\bar n}$ and ${\cal\bar{Y}} = \{ {\bar{\bf y}}_j\}_{j=1}^{\bar m}$ available in a different domain, which comprises cross-modal relationship $\mathcal{S} = \{ s_{ij}\}$, where $s_{ij} = 1$ implies points ${\bar{\bf x}}_i$ and ${\bar{\bf y}}_i$ are similar while $s_{ij} = 0$ indicates points ${\bar{\bf x}}_i$ and ${\bar{\bf y}}_i$ are disimilar. In real multimedia retrieval applications, the cross-modal relationship $S=\{s_{ij}\}$ can be collected from the relevance feedback information in click-through data, or from the social media where multiple modalities are usually presented.

The goal of Transitive Hashing Network (THN) is to learn two hash functions $f_x :\mathbb{R}^{d_x} \to \{ -1,1\}^b$ and $f_y :\mathbb{R}^{d_y} \to \{ -1,1 \}^b$ that encode data points from modalities $X$ and $Y$ into compact $b$-bit hash codes ${\bf h}_x = f_x({\bf x})$ and ${\bf h}_y = f_y({\bf y})$ respectively, such that the cross-modal relationship ${\cal S}$ can be preserved. With the learned hash functions, we can generate the hash codes ${\cal H}^q=\{{\bf h}_i^x\}_{i=1}^n$ and ${\cal H}^d=\{{\bf h}_j^y\}_{j=1}^m$ for the query modality and database modality respectively, which enables multimedia retrieval across heterogeneous data based on ranking the Hamming distances between hash codes.

To learn the transitive hash functions $f_x$ and $f_y$, we construct the training sets ${\cal X}=\{{\bf x}_i\}_{i=1}^{N}$ and ${\cal Y}=\{{\bf y}_j\}_{j=1}^{M}$ as follows: (1) ${\cal X}$ comprises the whole auxiliary dataset ${\bar{\cal X}}$ and another ${\hat n}$ data points randomly selected from the query set ${\cal X}^q$, where $N = {\bar n} + {\hat n}$; (2) ${\cal Y}$ comprises the whole auxiliary dataset ${\bar{\cal Y}}$ and another ${\hat m}$ data points randomly selected from the database set ${\cal Y}^d$, where $M = {\bar m} + {\hat m}$.

\begin{figure}[tbp]
  \centering
  \includegraphics[width=0.9\textwidth]{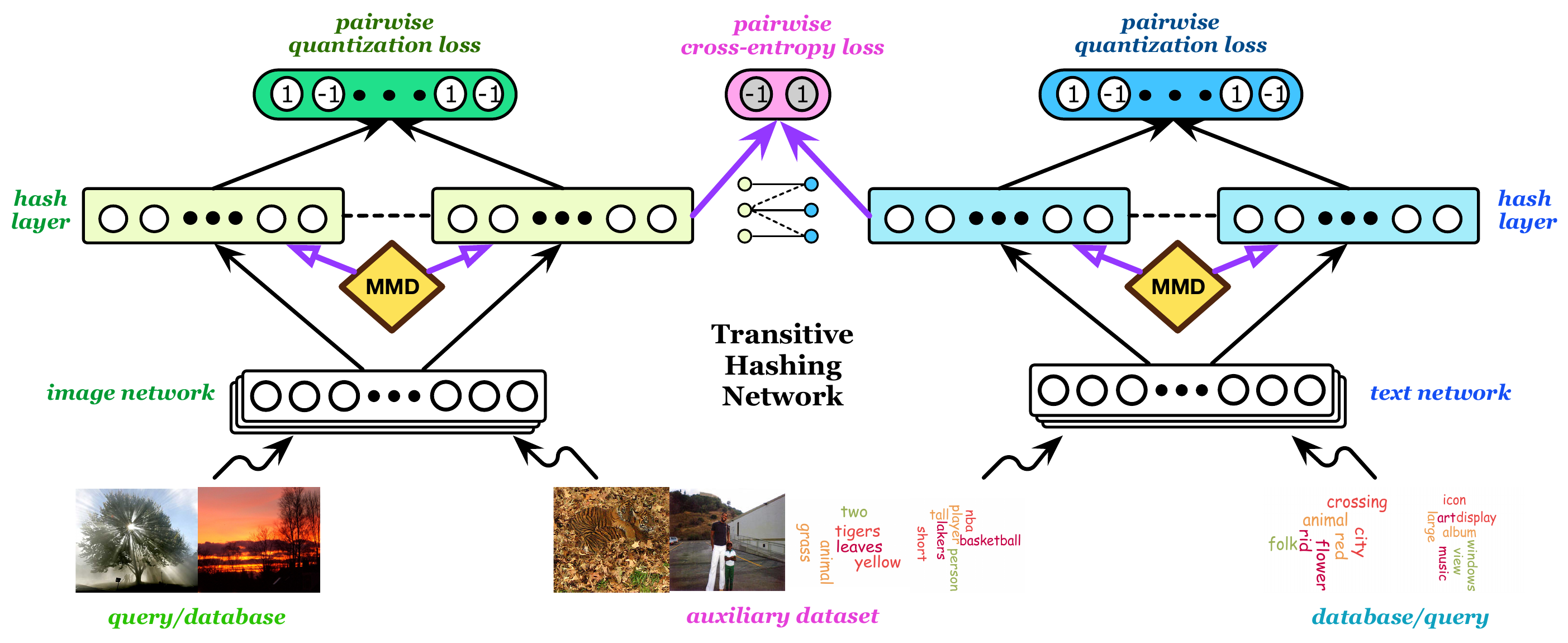}
  \caption{The hybrid architecture of Transitive Hashing Network (THN), which comprises heterogeneous relationship learning, homogeneous distribution alignment, and quantization error minimization. The key is to build a transitivity (in purple) from query to database across both modalities and domains.}
   \label{fig:THN}
\end{figure}

\subsection{Architecture for Transitive Hashing}
The architecture for learning transitive hash functions are illustrated in Figure~\ref{fig:THN}, which is a hybrid deep architecture comprising an image network and a text network. In the image network, we extend AlexNet \cite{cite:NIPS12CNN}, a deep convolutional neural network (CNN) comprised of five convolutional layers $conv1$--$conv5$ and three fully connected layers $fc6$--$fc8$. We replace the $fc8$ layer with a new $fch$ hash layer with $b$ hidden units, which transforms the network activation ${\bf z}_i^x$ in $b$-bit hash code by sign thresholding ${\bf h}^x_{i} = \operatorname{sgn} ({\bf z}_i^{x})$. In text network, we adopt the Multilayer perceptrons (MLP) \cite{cite:MIT86MLP} comprising three fully connected layers, of which the last layer is replaced with a new $fch$ hash layer with $b$ hidden units to transform the network activation ${\bf z}_i^y$ in $b$-bit hash code by sign thresholding ${\bf h}^y_{i} = \operatorname{sgn} ({\bf z}_i^{y})$. We adopt the hyperbolic tangent (tanh) function to squash the activations to be within $[-1,1]$, which reduces the gap between the $fch$-layer representation ${\bf z}_i^{\ast}$ and the binary hash codes ${\bf h}_i^{\ast}$, where $\ast\in\{x,y\}$. Several carefully-designed loss functions on the hash codes are added on top of the hybrid network for heterogeneous relationship learning and homogeneous distribution alignment, which enables query-database transitivity construction for heterogeneous multimedia retrieval.

\subsection{Heterogeneous Relationship Learning}
In this work, we jointly preserve the heterogeneous relationship ${\cal S}$ in the Hamming space and control the quantization error of sign thresholding in a Bayesian framework. We bridge the Hamming spaces of modalities $X$ and $Y$ by learning from the auxiliary dataset ${\bar{\cal X}}$ and ${\bar{\cal Y}}$. Note that, for a pair of binary codes ${\bf h}_i^x$ and ${\bf h}_j^y$, there exists a nice linear relationship between their Hamming distance $\mathrm{dist}_H(\cdot,\cdot)$ and inner product $\langle \cdot,\cdot \rangle$: ${\textrm{dis}}{{\text{t}}_H}\left( {{{\bf{h}}_i^x},{{\bf{h}}_j^y}} \right) = \frac{1}{2}\left( {K - \left\langle {{{\bf{h}}_i^x},{{\bf{h}}_j^y}} \right\rangle } \right)$. Hence in the sequel, we will use the inner product as a good surrogate of the Hamming distance to quantify the similarity between hash codes. Given heterogeneous relationship $\mathcal{S} = \{s_{ij}\}$, the logarithm Maximum a Posteriori (MAP) estimation of hash codes ${\bf H}^x = [{\bf h}_1^x,\ldots,{\bf h}_{\bar n}^x]$ and ${\bf H}^y = [{\bf h}_1^y,\ldots,{\bf h}_{\bar m}^y]$ can be defined as follows,
\begin{equation}\label{eqn:MAP}
	\begin{aligned}
  	\log p\left( {{\bf{H}^x, \bf{H}^y}|\mathcal{S}} \right) &\propto \log p\left( {\mathcal{S}|{\bf{H}^x, \bf{H}^y}} \right) p\left( {{{\bf{H}}^x}} \right)p\left( {{{\bf{H}}^y}} \right) \\
    &= \sum\limits_{{s_{ij}} \in \mathcal{S}} {\log p\left( {{s_{ij}}|{{\bf{h}}_i^x},{{\bf{h}}_j^y}} \right)p\left( {{{\bf{h}}_i^x}} \right)p\left( {{{\bf{h}}_j^y}} \right)} , \\
	\end{aligned}
\end{equation}
where  $p({\cal S}|{{\bf H}^x, {\bf H}^y})$ is the likelihood function, and $p({{\bf H}}^x)$ and $p({{\bf H}^y})$ are the prior distributions. For each pair, $p(s_{ij}|{\bf h}_i^x,{\bf h}_j^y)$ is the conditional probability of their relationship $s_{ij}$ given hash codes ${\bf h}_i^x$ and ${\bf h}_j^y$, which is defined as the pairwise logistic function,
 \begin{equation}
 \label{eqn:CDF}
	\begin{aligned}
	p\left( {{s_{ij}}|{{\bf h}_i^x},{{\bf h}_j^y}} \right) &= 
		\begin{cases}
			\sigma \left( {\left\langle {{s_{ij}}|{\bf h}_i^x,{\bf h}_j^y} \right\rangle } \right), & {s_{ij}} = 1 \\
			1 - \sigma \left( {\left\langle {{s_{ij}}|{\bf h}_i^x,{\bf h}_j^y} \right\rangle } \right), & {s_{ij}} = 0 \\
		\end{cases} \\
		& = \sigma {\left( {\left\langle {{s_{ij}}|{\bf h}_i^x,{\bf h}_j^y} \right\rangle } \right)^{{s_{ij}}}}{\left( {1 - \sigma \left( {\left\langle {{s_{ij}}|{\bf h}_i^x,{\bf h}_j^y} \right\rangle } \right)} \right)^{1 - {s_{ij}}}}, \\
	\end{aligned}
\end{equation}
where $\sigma \left( x \right) = {1}/({{1 + {e^{ - x}}}})$ is the sigmoid function and note that ${\bf h}_i^x = \operatorname{sgn}({{\bf z}_i^x})$ and ${\bf h}_i^y = \operatorname{sgn}({{\bf z}_i^y})$. Similar to logistic regression, the smaller the Hamming distance ${\text{dis}}{{\textrm{t}}_H}( {{{\bf{h}}_i^x},{{\bf{h}}_j^y}} )$ is, the larger the inner product ${\langle {{{\bf{h}}_i^x},{{\bf{h}}_j^y}} \rangle }$ will be, and the larger $p( {1|{{\bf{h}}_i^x},{{\bf{h}}_j^y}} )$ will be, implying that pair ${\bf h}_i^x$ and ${\bf h}_j^y$ should be classified as ``similar''; otherwise, the larger $p( {0|{{\bf{h}}_i^x},{{\bf{h}}_j^y}} )$ will be, implying that pair ${\bf h}_i^x$ and ${\bf h}_j^y$ should be classified as ``dissimilar''. Hence, Equation~\eqref{eqn:CDF} is a reasonable extension of the logistic regression classifier to the pairwise classification scenario, which is optimal for binary pairwise labels $s_{ij}\in\{0,1\}$. By MAP \eqref{eqn:CDF}, the heterogeneous relationship ${\cal S}$ can be preserved in the Hamming space.

Since discrete optimization of Equation~\eqref{eqn:MAP} with binary constraints ${\bf h}_i^\ast\in\{-1,1\}^b$ is difficult, for ease of optimization, continuous relaxation that ${\bf h}_i^x = {{\bf z}_i^x}$ and ${\bf h}_i^y = {{\bf z}_i^y}$ is applied to  the binary constraints, as widely adopted by existing hashing methods \cite{cite:Arxiv14Hashing}. To reduce the gap between the binary hash codes and continuous network activations, We adopt the hyperbolic tangent (tanh) function to squash the activations to be within $[-1,1]$. However, the continuous relaxation still gives rise to two issues: (1) uncontrollable quantization error by binarizing continuous activations to binary codes, and (2) large approximation error by adopting inner product between continuous activations as the surrogate of Hamming distance between binary codes. In this paper, to control the quantization error and close the gap between Hamming distance and its surrogate for learning accurate hash codes, we propose a new cross-entropy prior over the continuous activations $\{{\bf z}_i^\ast\}$, which is defined as follows,
\begin{equation}\label{eqn:prior}
  p\left( {{\mathbf{z}}_i^ * } \right) \propto \exp \left( { - \lambda H\left( {\frac{{\mathbf{1}}}{b},\frac{{\left| {{\mathbf{z}}_i^ * } \right|}}{b}} \right)} \right),
\end{equation}
where $\ast\in\{x,y\}$, and $\lambda$ is the parameter of the exponential distribution. We observe that maximizing this prior is reduced to minimizing the cross-entropy $H(\cdot,\cdot)$ between the uniform distribution ${{\mathbf{1}}/b}$ and the code distribution ${\left| {{\mathbf{z}}_i^ * } \right|/b}$, which is equivalent to assigning each bit of the continuous activations $\{ \bf{z}_i^\ast \}$ to binary values $\{ -1, 1 \}$.

By substituting Equations \eqref{eqn:CDF} and \eqref{eqn:prior} into the MAP estimation in Equation~\eqref{eqn:MAP}, we achieve the optimization problem for heterogeneous relationship learning as follows,
\begin{equation}\label{eqn:HRL}
\mathop {\min }\limits_\Theta  J = L + \lambda Q, \\
\end{equation}
where $\lambda$ is trade-off parameter between the pairwise cross-entropy loss $L$ and the pairwise quantization loss $Q$, and $\Theta  $ denotes the set of network parameters. Specifically, the pairwise cross-entropy loss $L$ is defined as
\begin{equation}\label{eqn:heteoL}
  {L} = \sum\limits_{s_{ij}\in{\cal S}} \log \left( {1 + \exp \left( {\left\langle {{\bf{z}}_i^x,{\bf{z}}_j^y} \right\rangle } \right)} \right) - {s_{ij}}\left\langle {{\bf{z}}_i^x,{\bf{z}}_j^y} \right\rangle. \\
\end{equation}
Similarly the pairwise quantization loss $Q$ can be derived as
\begin{equation}\label{eqn:heteQ}
  {Q} = \sum\limits_{s_{ij}\in{\cal S}} \sum\limits_{k = 1}^b (-\log ( {| {z_{ik}^x} |}) - \log ( {| {z_{jk}^y} |})).
\end{equation}

By optimizing the MAP estimation in Equation \eqref{eqn:HRL}, we can simultaneously preserve the heterogeneous relationship in training data and control the quantization error of binarizing continuous activations to binary codes. By learning from the auxiliary dataset, we can successfully bridge different modalities.

\subsection{Homogeneous Distribution Alignment}
The goal of transitive hashing is to perform efficient retrieval from the database of one modality in response to the query of another modality. Since there is no relationship between the query and the database, we exploit the auxiliary dataset ${\bar{\cal X}}$ and ${\bar{\cal Y}}$ to bridge the query modality and database modality. However, since the auxiliary dataset is obtained from a different domain, there are large distribution shifts between the auxiliary dataset and the query/database sets. Therefore, we should further reduce the distribution shifts by minimizing the Maximum Mean Discrepancy (MMD) \cite{cite:JMLR12MMD} between the auxiliary dataset and the query set (or between the auxiliary dataset and the database set) in the Hamming space. MMD is a nonparametric distance measure to compare different distributions $P_d$ and $P_x$ in reproducing kernel Hilbert space ${\cal H}$ (RKHS) endowed with feature map $\phi$ and kernel $k$ \cite{cite:JMLR12MMD}, formally defined as $D_q \triangleq \left\| {{\mathbb{E}_{{{\bf{h}}^q}\sim{P_q}}}\left[ {\phi \left( {{{\bf{h}}^q}} \right)} \right] - {\mathbb{E}_{{{\bf{h}}^x}\sim{P_x}}}\left[ {\phi \left( {{{\bf{h}}^x}} \right)} \right]} \right\|_\mathcal{H}^2$, where $P_q$ is the distribution of the query set ${\cal X}^q$, and $P_x$ is the distribution of the auxiliary set ${\bar{\cal X}}$. Using the same continuous relaxation, the MMD between the auxiliary dataset ${\bar{\cal X}}$ and the query set ${\cal X}^q$ can be computed as
\begin{equation}\label{eqn:HDA}
  {D_q} = \sum\limits_{i = 1}^{\hat n} {\sum\limits_{j = 1}^{\hat n} {\frac{{k\left( {{\bf{z}}_i^q,{\bf{z}}_j^q} \right)}}{{{\hat n^2}}}} }  + \sum\limits_{i = 1}^{\bar n} {\sum\limits_{j = 1}^{\bar n} {\frac{{k\left( {{\bf{z}}_i^x,{\bf{z}}_j^x} \right)}}{{{{\bar n}^2}}}} }  - 2\sum\limits_{i = 1}^{\hat n} {\sum\limits_{j = 1}^{\bar n} {\frac{{k\left( {{\bf{z}}_i^q,{\bf{z}}_j^x} \right)}}{{{\hat n}\bar n}}} },
\end{equation}
where $k(\bf{z}_i, \bf{z}_j) = \exp(-\gamma||\bf{z}_i-\bf{z}_j||^2)$ is the Gaussian kernel. Similarly, the MMD $D_d$ between the auxiliary dataset ${\bar{\cal Y}}$ and the query set ${\cal Y}^d$ can be computed by replacing the query modality with the database modality, i.e. by replacing $q$, $x$, $\hat n$ and ${\bar n}$ with $d$, $y$, $\hat m$, and ${\bar m}$ in Equation~\eqref{eqn:HDA}, respectively.

\subsection{Transitive Hash Function Learning}
To enable efficient retrieval from the database of one modality in response to the query of another modality, we construct the transitivity bridge between the query and the database (as shown by the purple arrows in Figure~\ref{fig:THN}) by integrating the objective functions of heterogeneous relationship learning \eqref{eqn:HRL} and the homogeneous distribution alignment \eqref{eqn:HDA} into a unified optimization problem as\begin{equation}\label{eqn:model}
  \mathop {\min }\limits_\Theta  C = J + \mu \left( {{D_q} + {D_d}} \right),
\end{equation}
where $\mu$ is a trade-off parameter between the MAP loss $J$ and the MMD penalty $(D_q+D_d)$. By optimizing the objective function in Equation~\eqref{eqn:model}, we can learn transitive hash codes which preserve the heterogeneous relationship and align the homogeneous distributions as well as control the quantization error of sign thresholding. Finally, we generate $b$-bit hash codes by sign thresholding as ${\bf h}^\ast = {\mathop{\rm sgn}} (\bf{z}^\ast)$, where  ${\mathop{\rm sgn}} (\bf{z})$ is the sign function on vectors that for each dimension $i$ of $\bf{z}^\ast$, $i=1,2,...,b$, ${\mathop{\rm sgn}} (z_i^\ast)=1$ if $z_i^\ast > 0$, otherwise ${\mathop{\rm sgn}} (z_i^\ast)=-1$. Since the quantization error in Equation \eqref{eqn:model} has been minimized, this final binarization step will incur small loss of retrieval quality.

We derive the learning algorithms for the THN model in Equation~\eqref{eqn:model} through the standard back-propagation (BP) algorithm. For clarity, we denote the point-wise cost with respect to ${\bar{\bf x}}_i$ as
\begin{equation}
\begin{aligned}
  {C_i} &= \textstyle{\sum\nolimits_{j:s_{ij}\in{\cal S}} {\log \left( {1 + \exp \left( {\left\langle {{\bf{z}}_i^x,{\bf{z}}_j^y} \right\rangle } \right)} \right) - {s_{ij}}\left\langle {{\bf{z}}_i^x,{\bf{z}}_j^y} \right\rangle } }  \\
   & - \textstyle{ \lambda \sum\limits_{j:s_{ij}\in{\cal S}} {\sum\limits_{k = 1}^b {\log (\left| {z_{ik}^x} \right|)} }  + \mu \sum\limits_{j = 1}^{\bar n} {\frac{{k\left( {{\bf{z}}_i^x,{\bf{z}}_j^x} \right)}}{{{{\bar n}^2}}}}  - \mu \sum\limits_{j = 1}^{\hat n} {\frac{{k\left( {{\bf{z}}_i^x,{\bf{z}}_j^q} \right)}}{{\hat n\bar n}}}}.  \\ 
\end{aligned} 
\end{equation}
In order to run the BP algorithm, we only need to compute the residual term $\frac{{\partial {C_i}}}{{\partial {{\tilde z}_{ik}}}}$, where ${{{\tilde z}_{ik}^x}}$ is the output of the last layer before activation function $a(\cdot)=\tanh(\cdot)$. We derive the residual term as
\begin{equation}\label{eqn:deltaC}
\begin{aligned}
  \frac{{\partial {C_i}}}{{\partial \tilde z_{ik}^x}} & = \textstyle{\sum\limits_{j:{s_{ij}} \in \mathcal{S}} {\left( {\left[ {\sigma \left( {\left\langle {{\bf{z}}_i^x,{\bf{z}}_j^y} \right\rangle } \right) - {s_{ij}}} \right]z_{jk}^y} \right)} a'\left( {\tilde z_{ik}^x} \right) - \frac{\lambda }{{z_{ik}^x}}\sum\limits_{j:{s_{ij}} \in \mathcal{S}} {a'\left( {\tilde z_{ik}^x} \right)} } \\
   & - \textstyle{ 2\mu \gamma \sum\limits_{j = 1}^{\bar n} {\frac{{k\left( {{\bf{z}}_i^x,{\bf{z}}_j^x} \right)}}{{{{\bar n}^2}}}\left( {z_{ik}^x - z_{jk}^x} \right)a'\left( {\tilde z_{ik}^x} \right)}  + 2\mu \gamma \sum\limits_{j = 1}^{\hat n} {\frac{{k\left( {{\bf{z}}_i^x,{\bf{z}}_j^q} \right)}}{{\hat n\bar n}}\left( {z_{ik}^x - z_{jk}^q} \right)a'\left( {\tilde z_{ik}^x} \right)}}.  \\ 
\end{aligned}
\end{equation}
The other residual terms with respect to modality $Y$ can be derived similarly. Since the only difference between standard BP and our algorithm is Equation~\eqref{eqn:deltaC}, we analyze the computational complexity based on Equation~\eqref{eqn:deltaC}. Denote the number of relationship pairs $\mathcal{S}$ available for training as $|\mathcal{S}|$, then it is easy to verify that the computational complexity is $O(|\mathcal{S}| + BN)$, where $B$ is mini-batch size.

\section{Experiments}\label{section:Experiments}

\subsection{Setup}
\textbf{NUS-WIDE}\footnote{\url{http://lms.comp.nus.edu.sg/research/NUS-WIDE.htm}} is a popular dataset for cross-modal retrieval, which contains 269,648 image-text pairs. The annotation for 81 semantic categories is provided for evaluation, which we prune by keeping the image-text pairs that belong to the 16 categories shared with ImageNet \cite{cite:IMAGENET}. Each image is resized into $256 \times 256$ pixels, and each text is represented by a bag-of-word (BoW) feature vector. We perform two types of cross-modal retrieval on the NUS-WIDE dataset: (1) using image query to retrieve texts (denoted by $I \rightarrow T$); (2) using text query to retrieve images (denoted by $T \rightarrow I$). The heterogeneous relationship ${\cal S}$ for training and the ground-truth for evaluation are defined as follows: if an image $i$ and a text $j$ (not necessarily from the same pair) share at least one of the 16 categories, they are relevant, i.e. relationship $s_{ij}=1$; otherwise, they are irrelevant, i.e. relationship $s_{ij}=0$.

\textbf{ImageNet-YahooQA} \cite{cite:KDD14HTH} is a heterogenous media dataset consisting of images from ImageNet \cite{cite:IMAGENET} and QAs from Yahoo Questions and Answers\footnote{\url{http://developer.yahoo.com/yql/}} (YahooQA). ImageNet is an image database of over 1 million images organized according to the WordNet hierarchy. We select the images that belong to the 16 categories shared with the NUS-WIDE dataset.  YahooQA is a text dataset of about 300,000 QAs crawled from a public API of Yahoo Query Language (YQL), detailed in \cite{cite:KDD14HTH}. Each QA is regarded as a text document and represented by a bag-of-word (BoW) feature vector. As the QAs are unlabeled, to enable evaluation, we assign one of the 16 category labels to each QA by checking whether the corresponding class word matches that QA. Note that, though the selected datasets from NUS-WIDE and ImageNet/YahooQA share the same set of labels, their data distributions are significantly different since they are collected from different domains. We perform two types of cross-modal retrieval on the ImageNet-YahooQA dataset: (1) using image query in ImageNet to retrieve texts from YahooQA (denoted by $I \rightarrow T$); (2) using text query in YahooQA to retrieve images from ImageNet (denoted by $T \rightarrow I$). The ground-truth for evaluation is consistent with that of the NUS-WIDE dataset.

We follow \cite{cite:KDD14HTH} to evaluate the retrieval quality based on standard evaluation metrics: Mean Average Precision (MAP) and Precision-Recall curves. We evaluate and compare the retrieval quality of the proposed \textbf{THN} approach with five state of the art cross-modal hashing methods, including two unsupervised methods Cross-View Hashing (\textbf{CVH}) \cite{cite:IJCAI11CVH} and Inter-Media Hashing (\textbf{IMH}) \cite{cite:SIGMOD13IMH}, two supervised methods Quantized Correlation Hashing (\textbf{QCH}) \cite{cite:IJCAI15QCH} and Heterogeneous Translated Hashing (\textbf{HTH}) \cite{cite:KDD14HTH}, and one deep hashing method Deep Cross-Modal Hashing (\textbf{DCMH}) \cite{cite:JDCMH16}. 

For fair comparison, all of the methods use identical training and test sets. For the deep learning based methods, including DCMH and the proposed THN, we directly use the image pixels as input. For the shallow learning based methods, we reduce the 4096-dimensional AlexNet features \cite{cite:ICML14DeCAF} of images to 500 dimensions using PCA, which incurs negligible loss of retrieval quality but significantly speeds up the evaluation process. For all methods, we use bag-of-word (BoW) features for text representations, which are reduced to 1000 dimensions using PCA for speeding up the evaluation.

We implement the THN model in \textbf{Caffe}. For image network, we adopt  AlexNet \cite{cite:NIPS12CNN}, fine-tune convolutional layer $conv1$--$conv5$ and fully-connected layer $fc6$--$fc7$ copied from the pre-trained model and train the $fch$ hash layer from scratch, all via back-propagation. Since $fch$ hash layer is trained from scratch, we set its learning rate to be $10$ times that of the other layers. For text network, we employ a three-layer MLP with the numbers of hidden units set to $1000$, $500$, and $b$, respectively. We use the mini-batch stochastic gradient descent (SGD) with $0.9$ momentum and the learning rate strategy in Caffe, cross-validate learning rate from $10^{-5}$ to $10^{-1}$ with a multiplicative step-size $10^{1/2}$. We train the image network and the text network jointly in the hybrid deep architecture by optimizing the objective function in Equation~\eqref{eqn:model}. The codes and configurations will be made available online.

\begin{table}[!htbp]
    \addtolength{\tabcolsep}{0.2pt} 
    \centering 
    \caption{MAP Comparison of Cross-Modal Retrieval Tasks on NUS-WIDE and ImageNet-YahooQA}
    \label{table:NusMAP}
    \begin{small}
    \vspace{-5pt}
    \begin{tabular}{c|c|cccc|cccc}
        \Xhline{1.0pt}
        \multirow{2}{20pt}{\centering Task} & \multirow{2}{20pt}{\centering Method} & \multicolumn{4}{c}{NUS-WIDE} & \multicolumn{4}{|c}{ImageNet-YahooQA}\\
        \cline{3-10}
        & & 8 bits & 16 bits  & 24 bits  & 32 bits & 8 bits & 16 bits  & 24 bits  & 32 bits \\
        \hline
        \multirow{6}{30pt}{\centering $ I \rightarrow T$} 
        & IMH \cite{cite:SIGMOD13IMH} & 0.5821 & 0.5794 &  0.5804 & 0.5776 & 0.0855 & 0.0686 &  0.0999 &  0.0889 \\
        & CVH \cite{cite:IJCAI11CVH} &  0.5681 &   0.5606  &  0.5451  &  0.5558 & 0.1229 & 0.1180 & 0.0941 & 0.0865\\
        & QCH \cite{cite:IJCAI15QCH}  &  0.6463  &   0.6921 &    0.7019  &  0.7127 & 0.2563  &  0.2494  &  0.2581  &  0.2590\\ 
        & HTH \cite{cite:KDD14HTH}  & 0.5232 & 0.5548 & 0.5684 & 0.5325 & 0.2931 & 0.2694 & 0.2847 & 0.2663 \\
        & DCMH \cite{cite:JDCMH16} & \underline{0.7887} & \underline{0.7397} & \underline{0.7210} & \underline{0.7460} & \underline{0.5133} & \underline{0.5109} & \underline{0.5321} & \underline{0.5087}\\
        \cline{2-10}
        & THN   & \textbf{0.8252}       & \textbf{0.8423} & \textbf{0.8495} & \textbf{0.8572}  & \textbf{0.5451} & \textbf{0.5507}  & \textbf{0.5803} & \textbf{0.5901} \\
        \hline
        \multirow{6}{30pt}{\centering $ T \rightarrow I$} 
        & IMH \cite{cite:SIGMOD13IMH} &  0.5579 & 0.5593 & 0.5528 & 0.5457 & 0.1105 & 0.1044 & 0.1183 & 0.0909\\
        & CVH \cite{cite:IJCAI11CVH} & 0.5261 &   0.5193 &   0.5097 &   0.5045 &0.0711 & 0.0728 & 0.1116 & 0.1008\\
        & QCH \cite{cite:IJCAI15QCH} & 0.6235  &  0.6609  & 0.6685 &  0.6773 &  0.2761  &   0.2847  &  0.2795  &  0.2665\\
        & HTH \cite{cite:KDD14HTH}  & 0.5603 & 0.5910 & 0.5798 & 0.5812 & 0.2172 & 0.1702 &  0.3122 &  0.2873\\
        & DCMH \cite{cite:JDCMH16} & \underline{0.7882} & \underline{0.7912} & \underline{0.7921} & \underline{0.7718} &\underline{0.5163} & \underline{0.5510} & \underline{0.5581} & \underline{0.5444} \\
        \cline{2-10}
& THN   & \textbf{0.7905}      & \textbf{0.8137} &    \textbf{0.8245}   & \textbf{0.8268} &\textbf{0.6032}  & \textbf{0.6097}& \textbf{0.6232} & \textbf{0.6102} \\
        \Xhline{1.0pt}
    \end{tabular}
    \end{small}
    \vspace{-10pt}
\end{table}

\begin{figure*}[htb]
    \centering
     \subfigure[$I\to T$ @ 24 bits]{
        \includegraphics[width=0.4\textwidth]{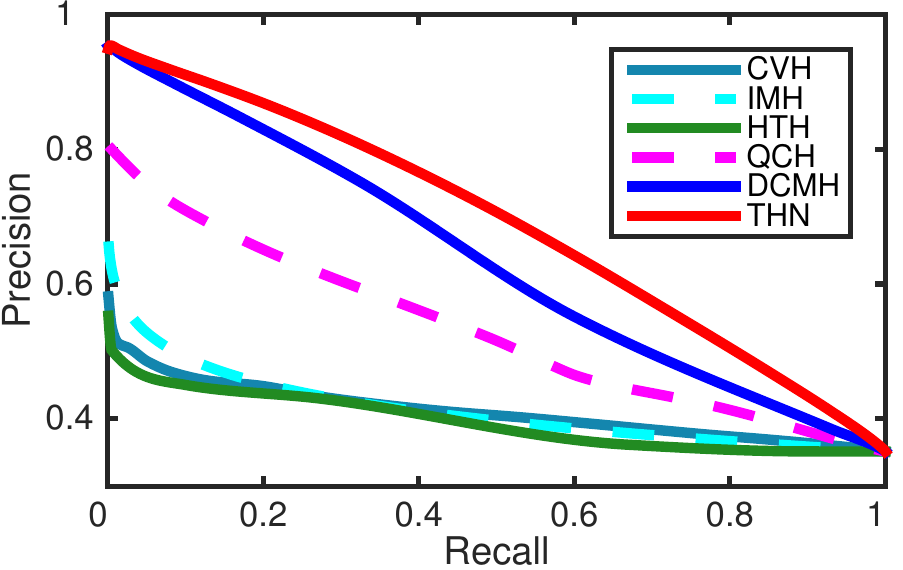}
        \label{fig:pr_nus_it}
    }
		\hfil
    \subfigure[$T\to I$ @ 24 bits]{
        \includegraphics[width=0.4\textwidth]{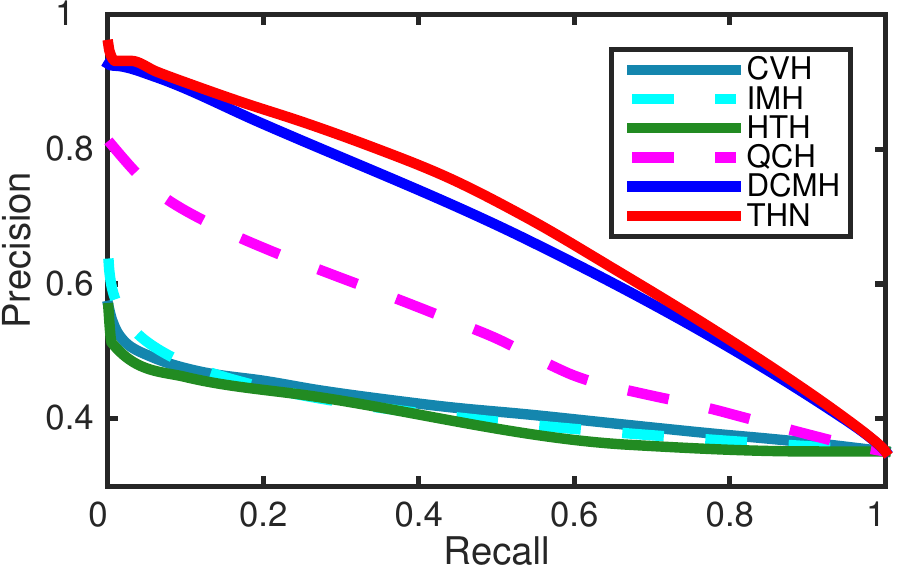}
        \label{fig:pr_nus_ti}
    }
    \vspace{-5pt}
    \caption{Precision-recall curves of Hamming ranking with 24-bits hash codes on NUS-WIDE.}
    \label{fig:pr_nus}
    \vspace{-20pt}
\end{figure*}

\subsection{Results}
\textbf{NUS-WIDE:} We follow the experimental protocols in \cite{cite:KDD14HTH}. We randomly select 2,000 images or texts as query set, and correspondingly, the remaining texts and images are used as the database. We randomly select 30 images and 30 texts per class distinctly from the database as the training set, which means that the images and texts are not paired so the relationship between them are heterogeneous.

We evaluate and compare the retrieval accuracies of the proposed THN with five state of the art hashing methods. The MAP results are presented in Table~\ref{table:NusMAP}. We can observe that THN generally outperforms the comparison methods on the two cross-modal tasks. In particular, compared to the state of the art deep hashing method DCMH, we achieve relative increases of \textbf{9.47\%} and \textbf{2.85\%} in average MAP for the two cross-modal retrieval tasks $I \to T$ and $T \to I$ respectively.

The precision-recall curves based on 24-bits hash codes for the two cross-modal retrieval tasks are illustrated in Figure~\ref{fig:pr_nus}. We can observe that THN achieves the highest precision at all recall levels. This results validate that THN is robust under diverse retrieval scenarios preferring either high precision or recall. The superior results in both MAP and precision-recall curves suggest that THN is a new state of the art method for the more conventional cross-modal retrieval problems where the relationship between query and database is available for training as in the NUS-WIDE dataset.

\textbf{ImageNet-YahooQA:} We follow similar protocols in \cite{cite:KDD14HTH}. We randomly select 2,000 images from ImageNet or 2000 texts from YahooQA as query set, and correspondingly, the remaining texts in YahooQA and the images in ImageNet are used as the database. For the training set, we randomly select 2000 NUS-WIDE images and 2000 NUS-WIDE texts as the supervised auxiliary dataset and select 500 ImageNet images and 500 Yahoo text documents as unsupervised training data. For all comparison methods, we note that they can only use the heterogeneous relationship in the supervised auxiliary dataset (NUS-WIDE) but cannot use the unsupervised training data from the query set and the database set (ImageNet and YahooQA). It is desirable that the THN model can use both supervised auxiliary dataset and unsupervised training data for heterogeneous multimedia retrieval.
\begin{table}[tp]
    \addtolength{\tabcolsep}{3.5pt} 
    \centering 
    \caption{MAP Comparison of Cross-Modal Retrieval Tasks of THN variants on ImageNet-YahooQA}
    \label{table:EmpirMAP}
    \small
    \vspace{-5pt}
    \begin{tabular}{c|cccc|cccc}
        \Xhline{1.0pt}
        \multirow{2}{20pt}{\centering Method} & \multicolumn{4}{c|}{\centering $ I \rightarrow T$} &\multicolumn{4}{c}{\centering $ T \rightarrow I$} \\
        \cline{2-9}
        & 8 bits & 16 bits  & 24 bits  & 32 bits & 8 bits & 16 bits  & 24 bits  & 32 bits\\
        \hline
         THN-ip &0.2976  & 0.3171 & 0.3302  & 0.3554 & 0.3443 & 0.3605  & 0.3852  & 0.4286 \\
         THN-D&\underline{0.5192}  & 0.5123 & 0.5312  & \underline{0.5411} & 0.5423 & 0.5512 & 0.5602 & 0.5489\\
         THN-Q & 0.4821& \underline{0.5213}  & \underline{0.5352}  & 0.4947 & \underline{0.5731}  & \underline{0.5592}& \underline{0.5849}  & \underline{0.5612}  \\
         THN & \textbf{0.5451} & \textbf{0.5507}  & \textbf{0.5803} & \textbf{0.5901} &\textbf{0.6032}  & \textbf{0.6097}& \textbf{0.6232} & \textbf{0.6102}  \\
        \Xhline{1.0pt}
    \end{tabular}
    \normalsize
    \vspace{-10pt}
\end{table}

\begin{figure}[!thp]
    \centering
     \subfigure[$I\to T$ @ 24 bits]{
        \includegraphics[width=0.4\textwidth]{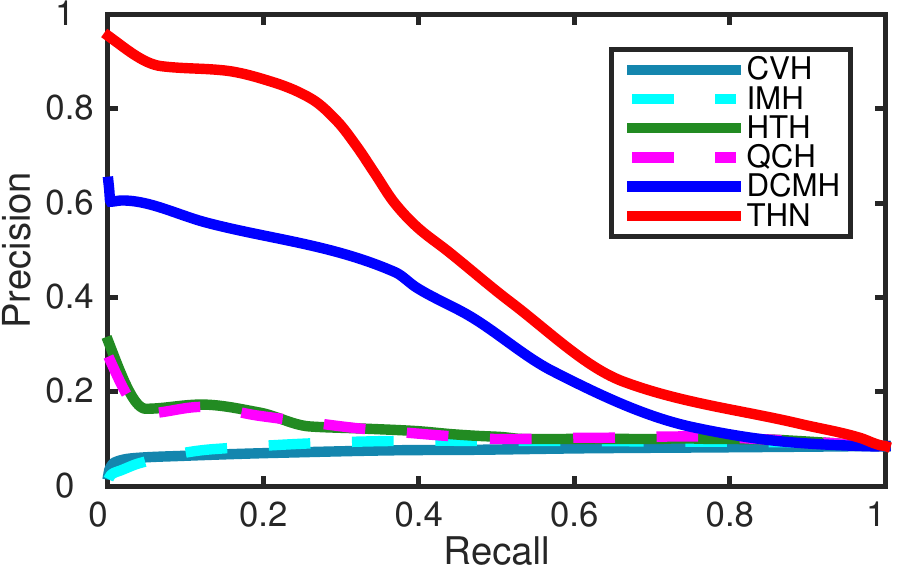}
        \label{fig:pr_yi_it}
    }
		\hfil
    \subfigure[$T\to I$ @ 24 bits]{
        \includegraphics[width=0.4\textwidth]{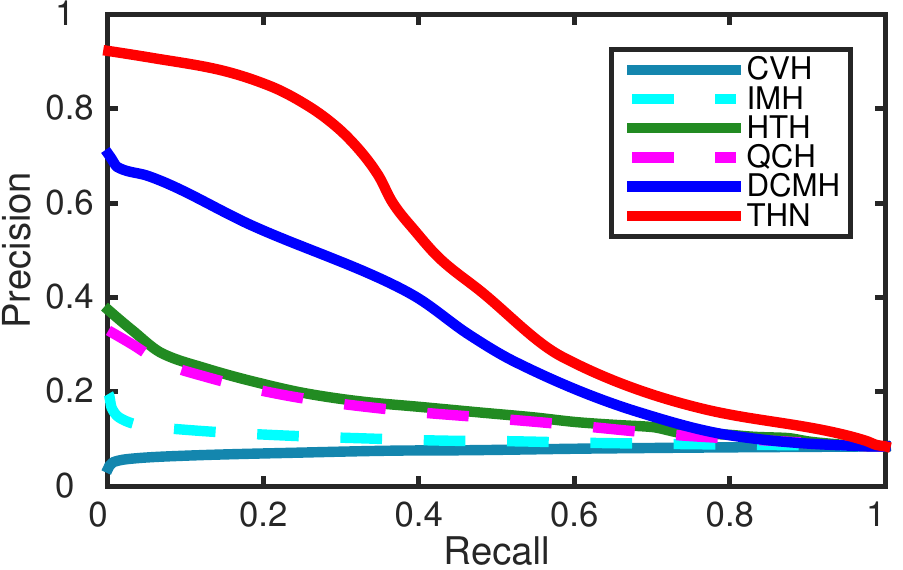}
        \label{fig:pr_yi_ti}
    }
    \vspace{-5pt}
    \caption{Precision-recall curves of Hamming ranking with 24-bits hash codes on Imagenet-YahooQA. }
    \label{fig:pr_iy}
    \vspace{-10pt}
\end{figure}

We evaluate and compare the retrieval accuracies of the proposed THN with five state of the art hashing methods. The MAP results are presented in Table~\ref{table:NusMAP}. We can observe that for these novel cross-modal and cross-domain retrieval tasks between ImageNet and YahooQA, THN outperforms the comparison methods on the two cross-modal tasks by very large margins. In particular, compared to the state of the art deep hashing method DCMH, we achieve relative increases of \textbf{5.03\%} and \textbf{6.91\%} in average MAP for the two cross-modal retrieval tasks $I \to T$ and $T \to I$ respectively. Similarly, the precision-recall curves based on 24-bits hash codes for the two cross-modal and cross-domain retrieval tasks in Figure~\ref{fig:pr_iy} shows that THN achieves the highest precision at all recall levels. 

The superior results in both MAP and precision-recall curves suggest that THN is a powerful approach to learning transitive hash codes, which enables heterogeneous multimedia retrieval between query and database across both modalities and domains. THN integrates heterogeneous relationship learning, homogeneous distribution alignment, and quantization error minimization into an end-to-end hybrid deep architecture for inferring the transitivity between query and database. The results on the NUS-WIDE dataset already shows that the heterogeneous relationship learning module is effective to bridge different modalities. The experiment on the ImageNet-YahooQA dataset further validates that the homogeneous distribution alignment between the auxiliary dataset and the query/database set, which is missing in all comparison methods, contributes significantly to the retrieval performance of THN. The reason is that the auxiliary dataset and the query/database sets are collected from different domains and follow different data distributions, hence there is substantial dataset shift which poses a major difficulty to bridge them. The homogeneous distribution alignment module of THN effectively close this shift by matching the corresponding data distributions with the maximum mean discrepancy. This makes the proposed THN model a good fit to heterogeneous multimedia retrieval problems.

\subsection{Discussion}
In order to study the effectiveness of THN, we investigate its variants on the ImageNet-YahooQA dataset: (1) \textbf{THN-ip} is the variant which uses the pairwise inner-product loss instead of the pairwise cross-entropy loss; (2) \textbf{THN-D} is the variant without using the unsupervised training data; (3) \textbf{THN-Q} is the variant without using the pairwise quantization loss. We report the MAP of all THN variants on ImageNet-Yahoo in Table~\ref{table:EmpirMAP}. We may have the following observations. (1) THN outperforms THN-ip by very large margins of 24.15\% / 23.19\% in absolute increase of average MAP, which confirms the importance of well-defined loss functions for heterogeneous relationship learning. (2) Compared to THN-D, THN achieves absolute increases of 4.06\% / 6.09\% in average MAP for the two cross-modal tasks $I \to T$ and $T \to I$. This convinces that THN can further exploit the unsupervised training data to bridge the Hamming spaces of auxiliary dataset (NUS-WIDE) and query/database sets (ImageNet-YahooQA), such that the auxiliary dataset can be served as a bridge to transfer knowledge between query and database. (3) THN also outperforms THN-Q by absolute promotions of 5.83\% / 4.20\% in average MAP, which confirms that the pairwise quantization loss can evidently reduce the quantization errors when binarizing the continuous representations to hash codes.

\section{Conclusion}
In this paper, we have formally defined a new transitive deep hashing problem for heterogeneous multimedia retrieval, and proposed a novel solution based on a hybrid deep architecture. The key to this problem is building the transitivity across different modalities and across different data distributions, which relies on relationship learning and distribution alignment. Extensive empirical evidence on public multimedia datasets show the proposed solution yields state of the art multimedia retrieval performance. In the future, we plan to extend the approach to online social media problems.

\medskip

\begin{small}

\end{small}

\end{document}